\documentclass[5p,times]{elsarticle}

\usepackage{multirow,bm,amsfonts,amsmath,booktabs}
\usepackage{hyperref}
\usepackage{color}

\DeclareMathOperator*{\argmax}{argmax} 

\journal{Journal of neurocomputing}

\begin{document}
	
	\begin{frontmatter}
		
		\title{An Attention-Based Word-Level Interaction Model: Relation Detection for Knowledge Base Question Answering}

		\author[mymainaddress,mysecondaryaddress]{Hongzhi Zhang}
		\ead{zhanghongzhi14@mails.ucas.ac.cn}
				
		\author[xuadd]{Guandong Xu}
		\ead{Guandong.Xu@uts.edu.au}
				
		\author[mymainaddress]{Xiao Liang}
		\ead{xliang@mail.ie.ac.cn}
		\author[mymainaddress]{Tinglei Huang\corref{mycorrespondingauthor}}
		\ead{tlhuang@mail.ie.ac.cn}
		
		\author[mymainaddress,mysecondaryaddress]{Kun Fu}
		\ead{kunfuiecas@gmail.com}
		
		\cortext[mycorrespondingauthor]{Corresponding author}
		
		\address[mymainaddress]{CAS Key Laboratory of Technology in Geo-spatial Information Processing and Application System, Institute of Electronics, Chinese Academy of Sciences, Beijing 100190, China}
		\address[mysecondaryaddress]{University of Chinese Academy of Sciences, Beijing 100049, China}
		\address[xuadd]{School of Software and Advanced Analytics, University of Technology Sydney, Australia}
		
		\begin{abstract}
			Relation detection plays a crucial role in Knowledge Base Question Answering (KBQA) because of the high variance of relation expression in the question.
			Traditional deep learning methods follow an encoding-comparing paradigm, where the question and the candidate relation are represented as vectors to compare their semantic similarity. 
			Max- or average- pooling operation, which compresses the sequence of words into fixed-dimensional vectors, becomes the bottleneck of information. 
			In this paper, we propose to learn attention-based word-level interactions between questions and relations to alleviate the bottleneck issue.
			Similar to the traditional models, the question and relation are firstly represented as sequences of vectors. 
			Then, instead of merging the sequence into a single vector with pooling operation, soft alignments between words from the question and the relation are learned.  
			The aligned words are subsequently compared with the convolutional neural network (CNN) and the comparison results are merged finally. 
			Through performing the comparison on low-level representations, the attention-based word-level interaction model (ABWIM) relieves the information loss issue caused by merging the sequence into a fixed-dimensional vector before the comparison. 
			The experimental results of relation detection on both SimpleQuestions and WebQuestions datasets show that ABWIM achieves state-of-the-art accuracy, demonstrating its effectiveness.
		\end{abstract}
		
		\begin{keyword}
			\texttt{relation detection; knowledge base question answering; word-level interaction; attention; semantic similarity}
		\end{keyword}
		
	\end{frontmatter}
	
	\section{Introduction }
	
	The prosperity of the web has greatly facilitated the dissemination of information. Large-scale knowledge bases (KB), such as Freebase and YAGO, have been built. Knowledge base question answering (KBQA) \cite{webqa-dataset, dongliMCNN} enables people to query the knowledge base with natural language and provides a feasible way of information acquisition. KBQA, which bridges the natural language and the structured knowledge base, involves two subtasks: determining the topic entity mentioned in the question and detecting the relation path from the topic entity to the answer, namely entity linking and relation detection \cite{improved}.
	Relation detection plays a vital role in KBQA and it is difficult because of the ambiguity and high variance of relation expression. As per reported, most of the wrong answers are caused by relation detection\cite{golubcharcnn}. 
	\begin{figure*}[!htb]
		\centering
		\includegraphics[width=0.8\textwidth]{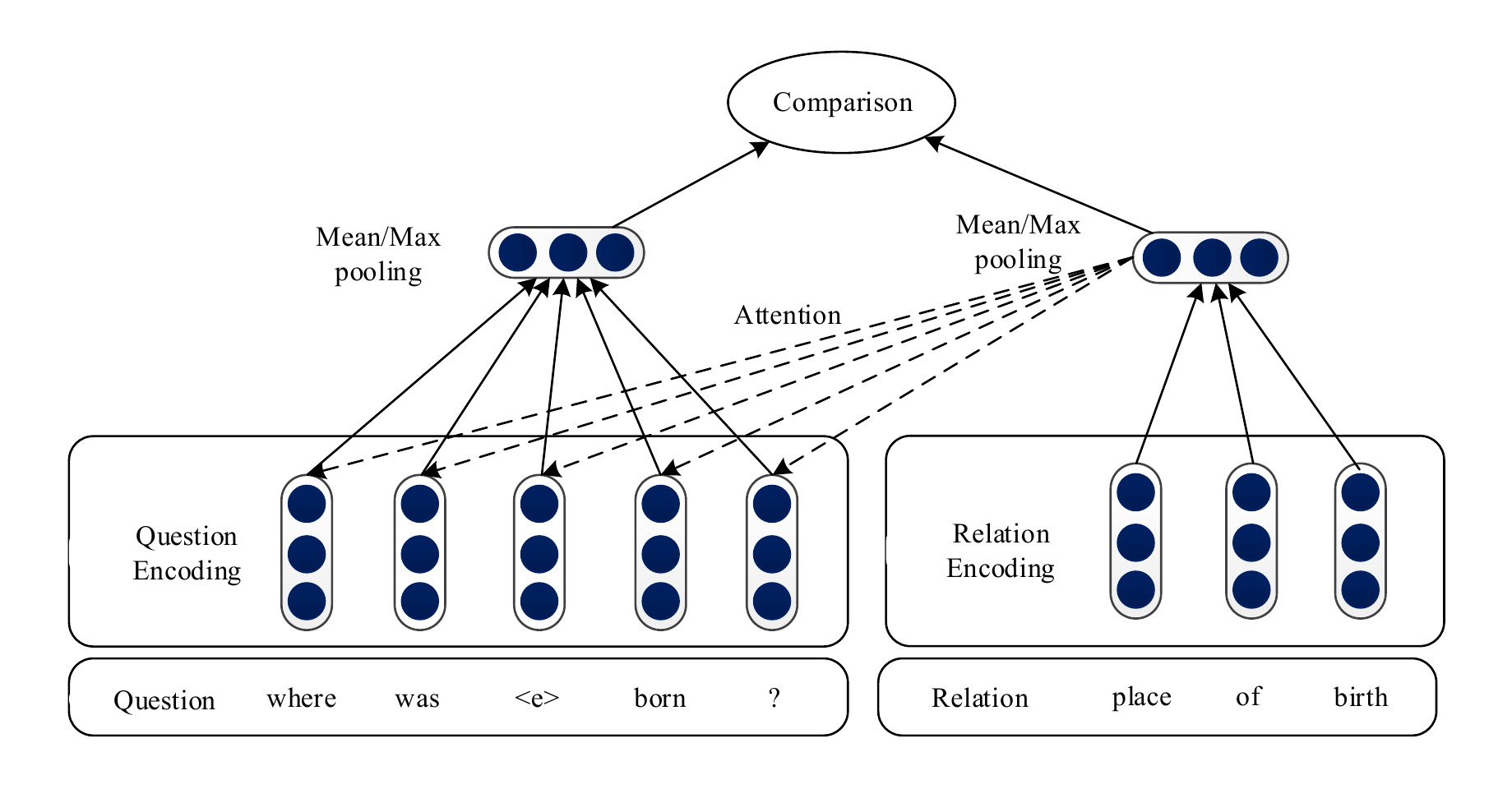}
		\caption{The encoding-comparing framework of KBQA relation detection model based on deep neural networks}
		\label{fig:traditioanlmethodfig}
	\end{figure*}
	
	Traditionally, relation detection for KBQA is mainly based on the semantic parsing \cite{semanticparsingbao, semanticparsinghe} method, where the natural language question is parsed into a logic query over the knowledge base. 
	These methods usually assume predefined lexical triggers, and suffer from lexical chasm \cite{lexicalchasm}. 
	Recently, deep learning methods are widely applied in the relation detection for KBQA. Most of deep learning methods fall into the encoding-comparing framework. As is illustrated in Figure \ref{fig:traditioanlmethodfig}, the deep neural networks firstly encode semantics of the question and the relation into vectors of the same dimension, and subsequently vector comparison is performed to measure their semantic similarity. 
	Many efforts have been made to the optimization of the relation and the question representation methods. 
	Relation representation methods mainly fall into two categories. 
	The first category takes the relation as a single semantic unit \cite{dai2016} and represents it with the pre-trained vector representation using network embedding methods such as TransE \cite{transE}. On the other hand, the other category \cite{BiCNN, golubcharcnn,AMPCNN,improved} takes the relation as a sequence of words or characters considering that the relation name comprises meaningful words. 
	Yu et al. \cite{improved} model relations from a combined perspective of the word-level and the holistic relation-level representations. 
	The question is regarded as a sequence of words, and different encoding methods are adopted to encode it from different granularities. 
	Specifically, Yih et al. \cite{BiCNN} model both questions as tri-grams of characters with CNN. He et al. \cite{golubcharcnn} process the questions as sequences of characters with an attention-based LSTM network. Hierarchical residual bi-directional LSTM (HR-Bi-LSTM) is proposed to learn the question representation with different levels of granularity \cite{improved}. 
	Attention mechanisms are incorporated to emphasize important units, and learn a relation-dependent representation of question, such as the attentive pooling method \cite{AMPCNN} and the attention-based LSTM network \cite{golubcharcnn}. 

	One issue of the encoding-comparing framework is the information loss during the encoding procedure. The question or the relation is firstly mapped into a sequence of word embedding vectors and CNN or LSTM layers are adopted to capture the context information. In order to merge the vector sequence into a single vector of fixed size, max-pooling, average-pooling or attention-weighted average-pooling is adopted. 
	The pooling procedure which compressing the question and the relation into fixed-dimensional vectors becomes the bottleneck of information. Crucial information for the semantic similarity measurement can be lost before the comparison. 

	There is a new trend to learn interactions between sequences at a low level in tasks of semantic similarity measurement, such as machine reading comprehension \cite{machinereading}, textual entailment \cite{textualentailment}, and answer selection \cite{compare-aggregate}. Competitive results have been reported on these tasks.
	
	This paper proposes to learn attention-based word-level interactions between question and relation to ameliorate the information loss problem. Rather than encoding the question and the relation into fixed-dimensional vector, we firstly compare semantics of words in the question with the relation and merge the comparison results to measure their semantic similarity. 	
	Specifically, the question and its candidate relations are represented as sequence of vectors with two bi-directional LSTM (Bi-LSTM) respectively.	Then we learn attention-based soft-alignments between the question words and relation words. A weighted sum of the relation vector sequence is calculated for every word in question.  
	Subsequently a multi-kernel CNN layer is involved to compare the question and relation at a fine granularity, and a max-pooling operation is performed to aggregate comparison results. Finally, a logistic layer scores the semantic similarity based on the extracted features. 
	With the question represented as a sequence of vectors and the comparison performed on the learned word interactions, information loss caused by the early merging can be avoided to some extent.
	
	The main contributions of this paper are summarized as follows:	
	\begin{itemize}
		\item We propose to learn word-level interactions between the question and the relation, and the attention mechanism is incorporated for a fine-grained alignment between words for relation detection of KBQA. 
		\item We propose a method to perform comparison on the word-level interactions and merge the comparison results. Specifically, a CNN layer is incorporated to compare semantic similarity at word-level, and the comparison results are merged with a max-pooling operation. 
		\item The proposed model achieves state-of-the-art results on the relation detection tasks \cite{improved} on SimpleQuestions \cite{simpleqadataset} and WebQuestions \cite{webqadataset} dataset, demonstrating the effectiveness of proposed method.
	\end{itemize}
	
	The rest of paper is organized as follows. Related work is reviewed in Section 2. The ABWIM is introduced in Section 3 in details. Experimental results and analysis are given in Section 4. Finally, we conclude the paper and discuss the future work in Section 5.
	
	\section{Related Work}
	This paper mainly involves work about the following three aspects: the traditional relation extraction, relation detection of KBQA, and sequence to sequence semantic similarity measurement.
	\subsection{Relation Extraction}
	As a key component of information extraction, relation extraction aims to determine whether the input text describes a type of relation between two corresponding entities. Assuming the relation belongs to a predefined relation set, relation extraction is modelled as a multi-class classification problem. Traditional methods \cite{traditionalRE1,traditionalRE2} focus on feature engineering. In recent years, many deep learning methods, from word embedding \cite{embRE1}, CNN \cite{liukangrelationextraction} to recursive neural network\cite{recursiveRE} and the attention based models \cite{attentionRE2, attentionRE1}, are adopted to extract features automatically. Deep learning methods rely heavily on the labelled data, and manual annotation is laborious and time-consuming. So distant supervision method is firstly introduced by \cite{distantsupervisionRE} to generate training data automatically, and many subsequent works \cite{distantRE1, distantRE2} focus on reducing the impact of labelling errors in the automatically generated training data.
	
	The relation detection of KBQA differs from the traditional relation extraction in two aspects. On the one hand, there are thousands of relations in the knowledge base, while there are only 74 relations in TAC-KBP2015 task, which has the largest predefined relation set. Some relations are even absent from the training data for the relation detection of KBQA, so it is inappropriate to formulate this task as a classification problem.
	On the other hand, entity information, such as entity embeddings or types, of the two entities used in the relation extraction are not available for the relation detection of KBQA, since there is only one argument (the topic entity) given. 
	\subsection{Relation detection for KBQA}
	Open domain KBQA task consists of two subtasks: identifying the topic entity mentioned in the question and determining the answer path (i.e. the path start from the topic entity to the answer node in the knowledge base). Determining the answer path is also the process of relation detection. 
	
	Considering thousands of relations in the open domain question answering tasks, relation detection is processed as a candidate-selection problem. Firstly, candidate relations are generated according to the result of entity linking, and then they are ranked by their semantic similarity with the question. Ranking methods consists of the traditional ones that rely on feature engineering and the deep learning ones that are capable of learning to extract features automatically. 
	Much work has been done on the extraction of discriminant features for relation extraction. Features for similarity measurement are extracted from four aspects, namely literal, derivation, synonym, and context, and learning-to-rank \cite{BastLRanking} is adopted for the relation detection. The dependency parsing tree \cite{yaoRD} is explored for relation detection. 
	Traditional methods cannot perform well when dealing with the high representational and linguistic variability between questions and relations. For example, the relation word of question ``where is $\langle e\rangle$ located?'' is ``located'', while the relation recorded in the KB is ``containedby''. In this case, traditional methods fail to capture the similarity because of the semantic chasm between symbols. 
	
	Deep learning method is firstly applied to the relation detection of KBQA in \cite{Bordes2014} and since then various models based on deep learning are developed. Most of these methods follow the encoding-comparing paradigm, which maps the question and the candidate relation to vectors respectively, and then computes the similarity between vectors as their semantic similarity.
	For example, Dai et al. \cite{dai2016} taken the relation as a single token and initialized with pre-trained vector learned by TransE \cite{transE}. Some work models the relation detection as a sequence matching and ranking problem, since the relation in KB is denoted by a sequence of meaningful words. Character-level representation is adopted in \cite{golubcharcnn}, to reduce the size of parameters and improve robustness in processing the out-of-vocabulary words. 
	A CNN model with attentive max-pooling (AMPCNN) is proposed in \cite{AMPCNN}. Traditional CNN layer is adopted to encode the question, and then a novel attentive max-pooling mechanism is proposed to summarize the question representation aiming at putting more weights on the relation description parts of the question. 
	In \cite{improved}, relations are represented using Bi-LSTM with inputs of different granularity (both on the word and the relation-level representations), and the question is represented by a hierarchical residual Bi-LSTM network with different abstract levels. 
	
	However, the aggregation operation such as max-pooling or average-pooling operation, which reduces the question from a sequence of vectors to a single vector, results in information loss before the comparison. The representation methods with attention mechanism concentrate on important units of the sequence. In this paper, comparison is performed before the aggregation operation at a lower level by learning word-level interactions, so as to avoid the information loss before comparison.
	\subsection{Sequence to sequence semantic similarity measurement}
	Sequence to sequence semantic similarity measurement is one of the fundamental tasks in natural language processing (NLP). It is involved in answer selection, natural language inference and machine reading comprehension. Under the traditional encoding-comparing paradigm, CNN \cite{CNNseq2seq} and recurrent neural network (RNN) \cite{LSTMseq2seq, LSTMseq2seq2} are exploited respectively, and attention-based methods \cite{ABCNNseq2seq, attentionSeq2seq} are further introduced to emphasize important units. To solve the problem that fixed-dimensional vector is insufficient to capture information in the input sequences, some efforts are made to the compare-aggregate framework. 
	For example, the matching-LSTM model \cite{textualentailment,compareaggSNIanother} and pairwise word interaction model \cite{pairwisesni} are proposed for the natural language inference task, and then the models are extended to machine reading comprehension task \cite{machinereading, multiPerspectiveMatchingReading}. State-of-the-art results are reported on multiple tasks simultaneously within the compare-aggregate framework \cite{compare-aggregate, multiPerspectiveMatchingSNIQA} .
	\def\hright{\stackrel{\rightarrow}{\bm{h}}}
	\def\hleft{\stackrel{\leftarrow}{\bm{h}}}
	
	\section{Method}
	\subsection{Task definition}
	Given a question, its topic entity (identified by entity linking) and candidate relations \(C=\left\{rel_1,rel_2,\dots,rel_{|C|}\right\}\) in knowledge base, relation detection aims to identify the relation (chain) mentioned in the question. That is, finding the chain of relations that connects the topic entity and the answer in the KB. 
	
	The relation detection task is formulated as a point-wise ranking problem. 
	For each relation \(r\) in the candidate relation set \(C\), the model computes its semantic similarity with the question \(s(q,r)\), and selects the relation path with the highest score as the answer path, formally:
	\begin{equation}r^+= \argmax_{r\in C}s(q,r)\end{equation}
	
	\begin{figure*}[!htb]
		\centering
		\includegraphics[width=0.8\textwidth]{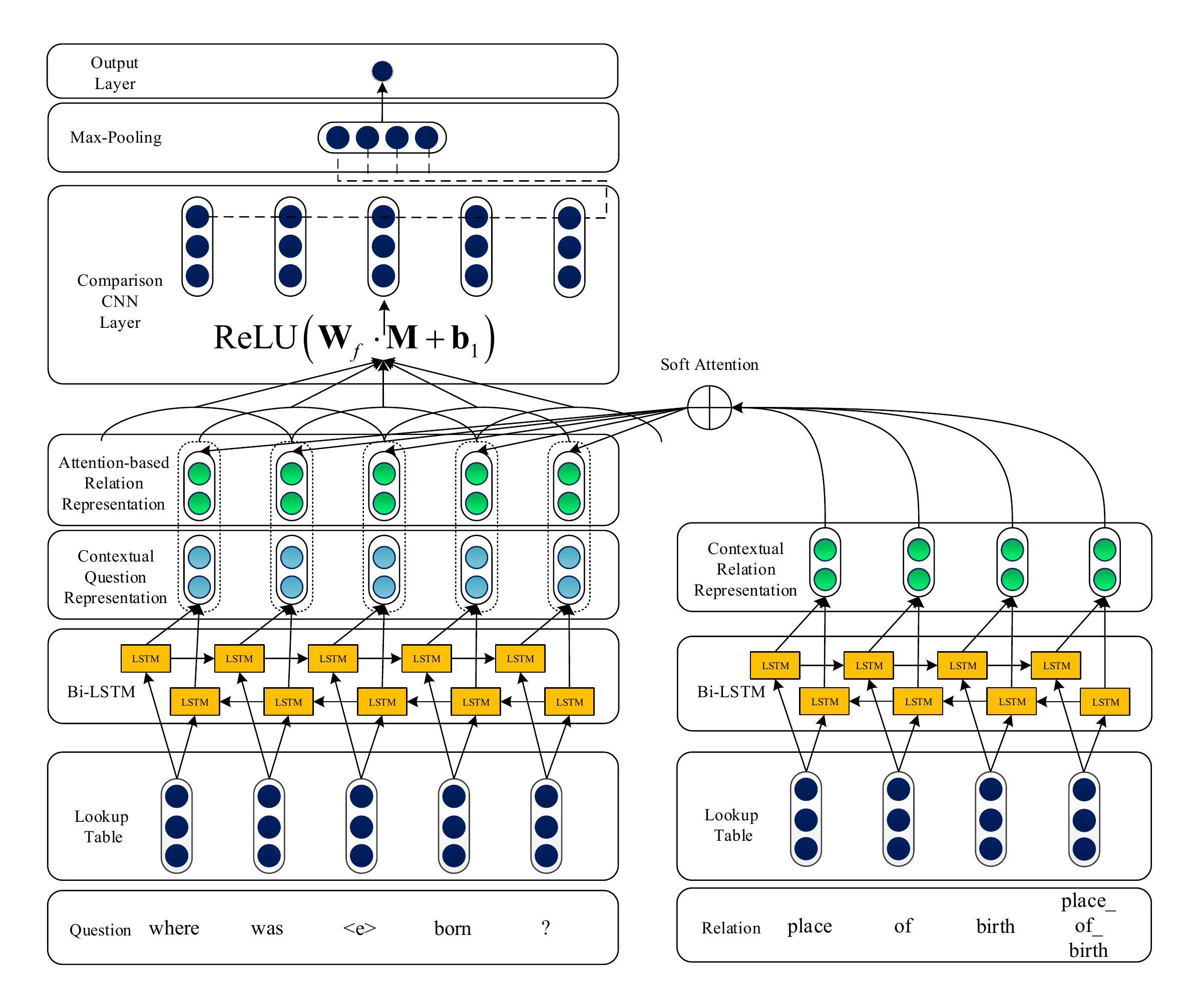}
		\caption{The framework of the attention-based word-level interaction model}
		\label{fig:framework}
	\end{figure*}
	
	\subsection{Model architecture}
	According to the above analysis, the key is to calculate the semantic similarity between the question and the candidate relation.
	The architecture of the model is given in Figure \ref{fig:framework}. The model consists of five components: 1) relation representation layer, which represents relation from diverse granularities with a Bi-LSTM layer. 2) question representation layer, another bi-LSTM is adopted to represent words in the question with both the past and future context in consideration. 3) word-level attention layer, that maps the relation representation to word representation of question with attention mechanism. 4) feature extraction layer, a CNN with multi-kennel \cite{kimcnn} is adopted to compare the question and mapped relation representation with context considered, then max-pooling operation is performed to aggregate the comparison results to extract the comparison features.
	5) output layer, a dense layer computes the final semantic similarity based on features extracted by the CNN layer and the max-pooling operation.
	
	\subsubsection{Relation representation layer}
	Both the relation and word-level representations \cite{improved} are adopted. 
	Relation level representation regards the relation, for example, ``place\_of\_birth'', as a single token, while word-level representation takes the relation as a sequence of words, i.e. ``place of birth''. 
	The relation-level representation focuses more on the global information of the relation such as the entity type of objective. 
	However, relation-level representation suffers from data sparsity because some relations are absent from the training data and their relation representation is initialized randomly during inference. 
	Through modelling the relation as a sequence of words, word-level representation relieves the impact of data sparsity. 
	Moreover, it enables us to perform word-level comparison between question and relation. 
	
	In this paper, a candidate relation (chain) \(r=\{r_1, r_{\left|r\right|}\}\), is represented as \(\{r_1^{word}, \dots, r_{\left|r\right|}^{word},\} \cup \{r_1^{rel}, r_{\left|r\right|}^{rel} \}\), where \(\left|r\right|\leq 2\) is the number of relations in the chain. 
	And we denote the number of tokens in the representation as \(|R|\).
	Therefore the combined representation of the input relation ``place\_of\_birth'' becomes \{``place'', ``of'', ``birth''\}\(\cup\)\{``place\_of\_birth''\}. Similarly, the input relation chain with two relations ``government\_positions\_held, from'' is represented as \(\{government, position, held, from\} \cup \{government\_positions\_held, from\}\). A lookup table transforms tokens in the sequence from ``one hot representation'' to corresponding embedding vectors of \(d\) dimensions. We have the word embedding vectors \(V\in \mathbb{R}^{|V|\times d}\), and the relation embedding vectors \(V_{rel}\in \mathbb{R}^{|V_{rel}|\times d}\), where \(|V|\) and \(|V_{rel}|\) are the vocabulary size and the number of relations in the KB respectively. 
	
\begin{figure}[h]
		\centering
		\includegraphics[width=0.45\textwidth]{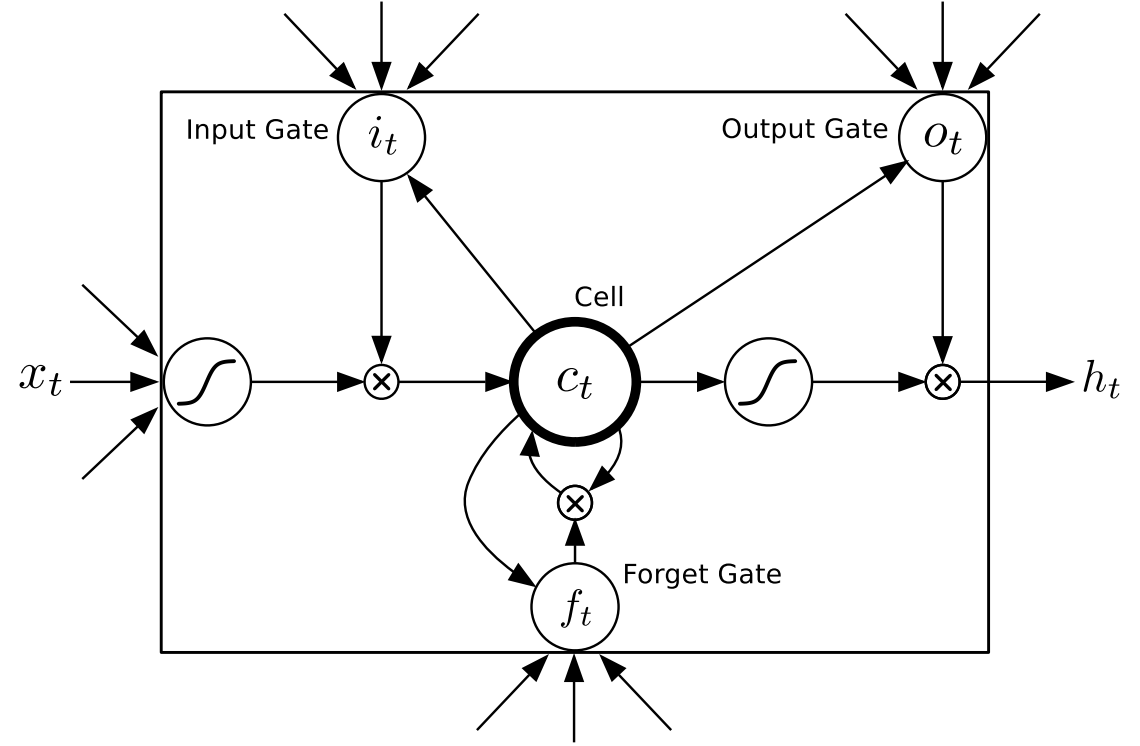}
		\caption{The structure of the LSTM cell}
		\label{fig:LSTM_cell}
	\end{figure}

	Then a Bi-LSTM layer is incorporated to represent word with its context in consideration.
	LSTM \cite{LSTM} is a variant of RNN. With the memory cell and the input, forget and output gate to manage the information flow, LSTM avoids the gradient exploding and vanishing problem and is capable of capturing the long range dependencies. The structure of the LSTM cell is illustrated in Figure \ref{fig:LSTM_cell}. 
	
	The forward LSTM cell outputs the encoding result based on the input \(\bm{x}_t\), the memory cell \(\bm{c}_{t-1}\) and the output of the last time \( \hright_{t-1}\). Here we denote the representation procedure in the cell as \(\hright_{t}=lstm(c_{t-1}, \hright_{t-1}, x_t)\). The three gates, namely input gate \(\bm{i}_t\), forget gate \(\bm{f}_t\) and output gate \(\bm{o}_t\), which respectively determine whether to update, reset and output values of the memory cell, are computed as follows
	\begin{equation}
	\bm{i}_t=\sigma\left(\bm{W}_i \bm{x}(t)+\bm{U}_i\hright_{t-1}+\bm{b}_i\right)
	\end{equation}
	\begin{equation}
	\bm{f}_t=\sigma\left(\bm{W}_f \bm{x}(t)+\bm{U}_f\hright_{t-1}+\bm{b}_f\right)
	\end{equation}
	\begin{equation}
	\bm{o}_t=\sigma\left(\bm{W}_o \bm{x}(t)+\bm{U}_o\hright_{t-1}+\bm{b}_o\right)
	\end{equation}
	where \(\bm{W}_*\in \mathbb{R}^{d_c\times d}\), \(\bm{U}_*\in\mathbb{R}^{d_c\times d_c}\) and \(\bm{b}*\in \mathbb{R}^{d_c}\) are parameters to be learned, hyper-parameter
	\(d_c\) is the dimension of LSTM cell and \(\sigma\) is the sigmoid function.
	The value of the memory cell is updated by
	\begin{equation}
	\bm{c}_t=\bm{i}_t\odot\tanh\left(\bm{W}_c \bm{x}(t)+\bm{U}_c\hright_{t-1}+\bm{b}_c\right)+\bm{f}_t\odot \bm{C}_{t-1}
	\end{equation}
	where \(\odot\) denotes element-wise multiplication, \(\bm{W}_c\in \mathbb{R}^{d_c\times d}\), \(\bm{U}_c\in \mathbb{R}^{d_c\times d}\) and \(\bm{b}_c\in \mathbb{R}^{d_c}\) are parameters to be learned. The output of the unit at time step \(t\) is
	\begin{equation}
	\hright_t=\bm{o}_t\odot\tanh(\bm{c}_t)
	\end{equation}
	Similarly, the backward LSTM cell is represented as
	\begin{equation}\hleft_t=lstm(\bm{c}_{t+1}, \hleft_{t+1}, \bm{x}_t)\end{equation}
	
	For input vector sequence \(\bm{X}=(\bm{x}_1,\bm{x}_2,\dots,\bm{x}_N)\) with length \(N\), forward LSTM encodes the input \(\bm{x}_t\) with context from \(\bm{x}_1\) to \(\bm{x}_{t-1}\) into vector \(\hright_t\), while backward LSTM encodes \(x_t\) to \(\hleft_t\) considering the future contextual information from \(\bm{x}_N\)  to \(\bm{x}_{t+1}\). 
	Concatenating \(\hright_t\) and \(\hleft_t\), the Bi-LSTM encodes the input \(\bm{x}_t\) with both the past and future information from the sentence in consideration. Then Bi-LSTM layer can be denoted by
	\begin{equation}\bm{H}=Bi-LSTM(\bm{X})=\left[ \left[\begin{array}{cc}
	\hright_{1}  \\
	\hleft_{1}   \end{array} \right],
	\left[ \begin{array}{cc}
	\hright_{2}  \\
	\hleft_{2}   \end{array} \right],
	\dots,
	\left[ \begin{array}{cc}
	\hright_{N}  \\
	\hleft_{N}   \end{array} \right] \right]
	\end{equation}
	
	In this paper, the context aware representation of relation can be formally defined as follows
	\begin{equation}
	\bm{R}=Bi-LSTM\left(\left[\bm{r}_1^{word}, \dots, \bm{r}_{|r|}^{word}, \bm{r}_1^{rel}, \dots, \bm{r}_{|r|}^{rel}  \right]\right)
	\end{equation}
	where \(\bm{R}\in \mathbb{R}^{2d_r\times|R|}\), \(d_r\) is the dimension of LSTM cell for the relation representation.
	
	\subsubsection{Question representation layer}
	Another Bi-LSTM layer is adopted to capture the context-aware representation of the question. For an input question \(q=\left\{w_1, w_2, \dots, w_{|Q|}\right\}\), where \(|Q|\) is the number of words in the question, every word \(w_i\) is mapped to its embedding vector \(\bm{w}_i\) by the lookup table. Therefore, the question is represented as a sequence of word vectors \(\bm{q}=\left\{\bm{w}_1, \bm{w}_2, \dots, \bm{w}_{|Q|}\right\}\). Note that words in questions and relations share the same word embedding. Finally, the context-dependent representation of these question words is defined as follows
	\begin{equation}\bm{Q}=Bi-LSTM(\left[\bm{w}_1, \bm{w}_2, \dots, \bm{w}_{|Q|}\right])\end{equation}
	where \(\bm{Q}\in \mathbb{R}^{2d_q\times |Q|}\), and \(d_q\) is the dimension of the LSTM cell. 
	\subsubsection{Attention layer}
	The attention layer aims to learn fine-grained interactions between the question and the relation. 
	For every word \(w_i\) in the question, the attention weight between \(w_i\) and the relation word \(w_j\) is calculated as follows
	\begin{equation}\alpha_{i,j}=\bm{r}_i^T \bm{W}_A \bm{q}_j\end{equation}
	where \(\bm{W}_A\in \mathbb{R}^{2d_r\times 2d_q}\) is a learnable matrix. 
	Then we have 
	\begin{equation}\bm{A}=softmax\left(\bm{R}^T \bm{W}_A \bm{Q}\right)\end{equation} 
	where \(\bm{A}\in \mathbb{R}^{|R|\times |Q|}\) is the attention weights between the words in the question and the relation, i.e.
	\begin{equation}
	a_{i,j}=\frac{exp(\alpha_{i,j})}{\sum_{k=1}^{|Q|}\sum_{l=1}^{|R|}exp(\alpha_{k,l})}
	\end{equation} 
	Then a weighted sum of relation vectors is calculated with the attention to word \(w_i\),
	\begin{equation}\widehat{\bm{r}}_i=\sum_{k=1}^{|R|}a_{i,k} \bm{r}_k\end{equation}
	It can be seen that \(\widehat{\bm{r}}_i\) encodes parts of the relation that best matches the question word \(w_i\). 
	Then overall we have
	\begin{equation}\widehat{\bm{R}}=\bm{R}\bm{A}\end{equation}
	and \(\widehat{\bm{R}}\in \mathbb{R}^{2d_r\times |Q|}\).
	
	By concatenating the representation of question words and the corresponding weighted representation of the relation, we get the word interactions \(\bm{M}\) between the question and the relation 
	\begin{equation}\bm{M}=\left[\left[ \begin{array}{cc}
	\bm{q}_{1}  \\
	\widehat{\bm{r}}_{1}   \end{array} \right],
	\left[ \begin{array}{cc}
	\bm{q}_{2}  \\
	\widehat{\bm{r}}_{2}   \end{array} \right],
	\dots,
	\left[\begin{array}{cc}
	\bm{q}_{N}  \\
	\widehat{\bm{r}}_{N}   \end{array} \right]\right]\end{equation}
	where \(\bm{M} \in \mathbb{R}^{2(d_q+d_r)\times |Q|}\). 
	
	\subsubsection{Feature extraction layer}
	Instead of the explicit element-wise comparison \cite{compare-aggregate}, a CNN layer with multiple kernel sizes \cite{kimcnn} is adopted to compare the semantics of the question and the relation with different context scopes. 
	Then the comparison results are aggregated with the max-pooling operation to extract the most discriminant comparison results as features.
	
	Let \(\bm{m}_j\in \mathbb{R}^{2(d_q+d_r)}\) be the \(j\)-th column of matrix \(\bm{M}\), and \(\bm{m}_{i:i+j}=\bm{m}_i\oplus \bm{m}_{i+1}\oplus \dots \oplus \bm{m}_{i+j}\), \(\bm{m}_{i+j}, \bm{m}_{i:i+j} \in \mathbb{R}^{2(d_q+d_r)}\) represents the concatenation of columns \(\bm{m}_i, \bm{m}_{i+1},\dots,\bm{m}_{i+j}\), where \(\oplus\) is the column concatenation operator. Then the convolution operation involves a filter with kernel size of \(k\), i.e. \(\bm{w}\in \mathbb{R}^{2(d_q+d_r)\times k}\), is performed to generate a feature. For example, a feature \(c_i\) is generated by applying the filter at \(\bm{m}_{i:i+k}\)
	\begin{equation}
	c_i=relu\left(\bm{w}\cdot \bm{m}_{i:i+k}\right)
	\end{equation}
	where \(relu\) is the rectified linear unit \cite{relu} activation.
	By applying the filter at every time step, we get the features along with the sequence
	\begin{equation}\bm{c}=\left[c_1, c_2, \dots, c_{|Q|}\right]\end{equation}
	After that, a max-over-time pooling operation is performed, yielding \(\widehat{c}=max(\bm{c})\) as the feature captured by this filter. The filter can be seen as a pattern matcher, to detect whether there is \(k\)-gram pattern in the sequence regardless its position. Thus this is a procedure of automatic feature extraction. Aiming at the semantic similarity measurement, the filter learns to capture features relevant to the semantic similarity calculation, and the max-pooling operation selects the significant one along the sequence. We adopt \(d_f\) filters thus getting \(d_f\) dimensional semantic similarity features \(\bm{f} \in \mathbb{R}^{d_f} \).
	Kernels with different sizes are used to capture comparison patterns with multiple lengths.
	
	The CNN layer conducts a fine-grained comparison between the question and the relation, and the max-pooling serves as a primary aggregation operation. With the comparison performed at a low-level representation, our model relieves the information loss problem caused by aggregating the information into a fixed-dimensional vector before comparison.
	\subsubsection{Output layer}
	Given the features \(\bm{f}\) extracted by the CNN layer and max-pooling, the output layer calculates the semantic similarity between the question and the relation.
	\begin{equation}o=\bm{w}_o^T\bm{f}\end{equation}
	where \(\bm{w}_o \in \mathbb{R}^{d_f}\) are parameters to be learned.
	\subsection{Training and inference}
	Training aims to give the golden relation higher score than the false relation. Pairwise ranking loss is defined as
	\begin{equation}
	L=\sum_{q_i \in \mathbb{Q}}\sum_{j\in C_{q_i}}\sigma\left(S(q_i, r_j^-|\theta)-S(q_i,r^+_i|\theta)\right)
	\end{equation}
	where \(\mathbb{Q}\) denotes the questions in the training set, \(C_{q_i}\) is the false candidate relation set, \(r^+_i\) is the golden relation of question \(q_i\), \(S\) denotes ABWIM. \(\theta\) denotes parameters of the network, and it consists of word embeddings, relation embeddings, parameters in the LSTM network for relation and question representation, feature maps in the CNN network and \(\bm{w}_o\) in the output layer.
	
	Backpropagation method is adopted to update the parameters. Formally, the parameters in \(\theta\) are updated by 
	\begin{equation}\theta=\theta-\lambda\frac{\partial L}{\partial \theta}\end{equation}
	where \(\lambda\) is the learning rate. Adadelta optimizer \cite{adadelta} is adopted to adjust the learning rate.
	Dropout is added to the output of embedding, LSTM, and CNN layer so as to avoid the overfitting problem.
	
	During inference, the semantic similarity \(S(q,r|\theta)\) is calculated for each candidate relation \(r\in C_q\), and the relation with highest semantic similarity score is regarded as the corresponding relation.
	\section{Experiment}
	\subsection{Experimental setting}
	\subsubsection{Datasets and metrics}
	To evaluate the effectiveness of ABWIM, experiments are carried out on the relation detection of two benchmark KBQA tasks, namely SimpleQuestions \cite{simpleqadataset} and WebQuestions \cite{webqadataset}.
	SimpleQuestions is a single-relation question-answering dataset, and its KB is a subset of Freebase and contains 2M entities. 
	Questions in this dataset can be answered by referring to single triple from the KB, so the length of relation chain \(|r|=1\). 
	WebQuestions is a multi-relation question-answering dataset, and the whole Freebase is used as its KB. 
	Similar procedure is adopted to generate the relation detection datasets based on these KBQA datasets. 
	Firstly, entity linking is performed to extract the topic entity from the question. 
	Then relations and relation chains (only on WebQuestions dataset and chain length no greater than 2) connected to the topic entity are taken as candidate relations. 
	After that, labelling the relation which points to the answer node as positive samples and other relations as negative samples. 
	Finally, the mention of the topic entity in the question is replaced with a special token \(\langle e\rangle \).
	Yin et al. \cite{AMPCNN} and Yu et al. \cite{improved} released their relation detection datasets\footnote{The datasets are available at https://github.com/Gorov/KBQA\_RE\_data} based on SimpleQuestions and WebQuestions, and we also use the datasets for fair comparison.
	
	Accuracy metric is used to measure the relation detection result,
	\begin{equation}
	A=\frac{N_c}{N}
	\end{equation}
	where N is the number of questions in the dataset, and \(N_c\) is the number of questions whose relation is correctly identified. 
	\subsubsection{Hyper parameters}
	The hyper-parameters of ABWIM are summarized in Table \ref{table: experimentconfig}. 
	In the experiment, word embeddings are initialized with the GloVe \cite{glove} with \(d=300\), and embeddings of relations and words that are out of vocabulary are randomly initialized by sampling values uniformly from (-0.25, 0.25). 
	Values of embedding are updated during the training process.
	
	All the experiments are carried out on a machine with a Nvidia GTX1080 GPU, and neural networks are implemented in Keras\footnote{https://github.com/fchollet/keras} with Tensorflow\footnote{https://github.com/tensorflow/tensorflow} as the backend.
	\begin{table}[!htb]
		\centering
		\caption{Hyper parameters of the ABWIM. (SQ and WQ denote the SimpleQustion and WebQuestions datasets respectively)}
		\label{table: experimentconfig}
		\begin{small}
			\begin{tabular}{cccc}
				\hline
				Parameter&Search Space&SQ&WQ\\
				\hline
				Embedding dim. \(d\) & 300 & 300 & 300\\
				Dim. of LSTM \(d_q,d_r\)&	50, 100, 150, 200 & 150 &10{\tiny }0\\
				CNN kernel size & [3], [1,3,5] & [1,3,5] & [1,3,5]\\
				num. of filters & 50, 100, 150, 300 & 150 & 100 \\
				Dropout rate & 0.2, 0.3, 0.35, 0.4, 0.5	& 0.35 &	0.35 \\
				Batch size&	64, 128, 256, 512&256 & 128 \\
				\hline
			\end{tabular}
		\end{small}
	\end{table}
	
	\subsubsection{Baselines}
	We compare the ABWIM with several baselines. 
	All the baseline models fall into the encoding-comparing paradigm, which firstly map the question 
	and relation as vectors and then get the semantic similarity by vector comparison. 
	These methods vary in the representation procedure, specifically:
	\begin{itemize}
		\item Bi-LSTM: two Bi-LSTM layers represent the question and the relation respectively, and the output of the last timestep is taken as their vector representation.
		\item BiCNN \cite{BiCNN}: both the question and the relation are represented by CNN with the word hash trick on letter-tri-grams.
		\item AMPCNN \cite{AMPCNN}, namely the CNN with attentive the max-pooling. 
		CNN is adopted to represent the question and the relation respectively. 
		An attentive max-pooling operation is performed to get the vector representation of question, aiming at putting more weights on the words that indicate the relation.
		\item HR-Bi-LSTM (Hierarchical Residual Bidirectional LSTM model) \cite{improved}, word and relation-level representations of relation are adopted, and the question is represented with HR-Bi-LSTM to obtain hierarchies of abstraction. 
	\end{itemize}
	\subsection{Result}
	\begin{table}[!htb]
		\centering
		\caption{Comparison of accuracy with the baselines on the SimpleQuestions and WebQuestions}
		\label{table:res0}
		\begin{small}
			\begin{tabular}{cccc}
				\hline
				Model & Relation Input & SimpleQuestions & WebQuestions \\
				\hline
				APMCNN & Words & 91.3 & -\\
				BiCNN & Words & 90.0 & 77.74 \\
				Bi-LSTM & Words & 91.2 & 79.32 \\
				HR-Bi-LSTM & Words + relations & 93.3 & 82.53\\
				Our method & Words + relations & \textbf{93.5} & \textbf{85.32} \\
				\hline
			\end{tabular}
		\end{small}
	\end{table}
	The relation detection results are shown in Table \ref{table:res0}. Our method  comes close to the performance of the state-of-the-art method on SimpleQuestions dataset. On the WebQuestions dataset, our method outperforms the best baseline methods by 2.79 points.  
	Questions and relations in the WebQuestion dataset are longer and the information loss problem is more serious, thus ABWIM outperforms the encoding-comparing baseline methods. 
	The experimental results demonstrate the effectiveness of ABWIM, which learns the word-level interactions and aggregates the comparison results with multi-kernel CNN networks.
	
	\subsection{Model Ablation Analysis}
	Ablation experiments are further carried out to analyse the effectiveness of the attention-based word-level interactions and the sequence preprocessing layer. 
	
	\begin{table*}[!htb]
		\centering
		\caption{Ablation experiment results}
		\label{tab: abaltion_res}
		\begin{small}
			\begin{tabular}{cccc}
				\hline
				Analysis content & Model & SimpleQA & WebQA \\
				\hline
				\multirow{3}{*}{Attention and framework}
				&Encoding-comparing model& 92.8 & 78.41 \\
				&Encoding-comparing model with bi-directional attention& 93.2 & 84.44 \\ 
				&Word-level interaction model without attention & 93.0 & 83.81\\
				\hline
				\multirow{3}{*}{Context representation }
				&Without preprocessing layer &92.9 &82.49\\
				&Gated liner transformation&92.9 &84.71\\
				& Fully CNN network & 93.3 & 84.79\\
				\hline
				-&ABWIM (our method)&\textbf{93.5} & \textbf{85.32} \\
				\hline
			\end{tabular}
		\end{small}
	\end{table*}
	\subsubsection{Attention and word-level interaction mechanisms}
	Ablation experiments for the attention and word-level interaction mechanisms are as follows. 
	\begin{itemize}
		\item Encoding-comparing model: questions and relations are represented with Bi-LSTM layers and the final output are taken as their representations. 
		Then semantic similarity is measured by the cosine similarity between vectors, formally
		\begin{equation}
		o=\cos(\bm{q}, \bm{r})
		\label{eq:cossim}
		\end{equation}
		
		\item Encoding-comparing model with bi-directional attention: following the traditional representation-comparison schema, questions and relations are firstly represented as vectors. 
		Bi-directional LSTM layer is adopted to perform the context aware representation, and the representation results are merged to vectors with attention mechanism. 
		Attention weights for questions are calculated by column wise max-pooling,
		\begin{equation}
		\widehat{a}_i = \max\left(a_{i,1}, a_{i,2},\dots, a_{i, |R|}\right)
		\end{equation}
		and soft-max operation,
		\begin{equation}
		\alpha_i = \frac{e^{\widehat{a}_i}}{\sum_{j=1}^{|Q|} e^{\widehat{a}_j}}
		\end{equation}
		Then the vector representation of the question is
		\begin{equation}
		\widehat{\bm{q}} =\sum_{i=1}^{|Q|}\alpha_i \bm{q}_i
		\end{equation}
		where \(\bm{q}_i\) is the \(i\)-th column of \(\bm{Q}\). Similarly, attention weights of the relation are calculated by
		\begin{equation}
		\widehat{b}_i = \max\left(a_{1,i}, a_{2,i},\dots, a_{|Q|,i}\right)
		\end{equation}
		and
		\begin{equation}
		\beta_i = \frac{e^{\widehat{b}_i}}{\sum_{j=1}^{|R|} e^{\widehat{b}_i}}
		\end{equation}
		Then relation's vector representation is
		\begin{equation}
		\widehat{\bm{r}} =\sum_{i=1}^{|A|}\beta_i \bm{r}_i
		\end{equation}
		Finally, the semantic similarity is calculated as
		\begin{equation}
		o=\cos(\widehat{\bm{q}}, \widehat{\bm{r}})
		\end{equation}
		\item Word-level interaction model without attention: removing the attention mechanism from the ABWIM, i.e. the attention weights are set equally. Formally, all the elements in \(\bm{A}\in \mathbb{R}^{|R|\times |Q|}\) are set to \(\frac{1}{|R|\times |Q|}\).
	\end{itemize}
	
	\begin{figure}[!htb]
		\centering
		\includegraphics[width=0.45\textwidth]{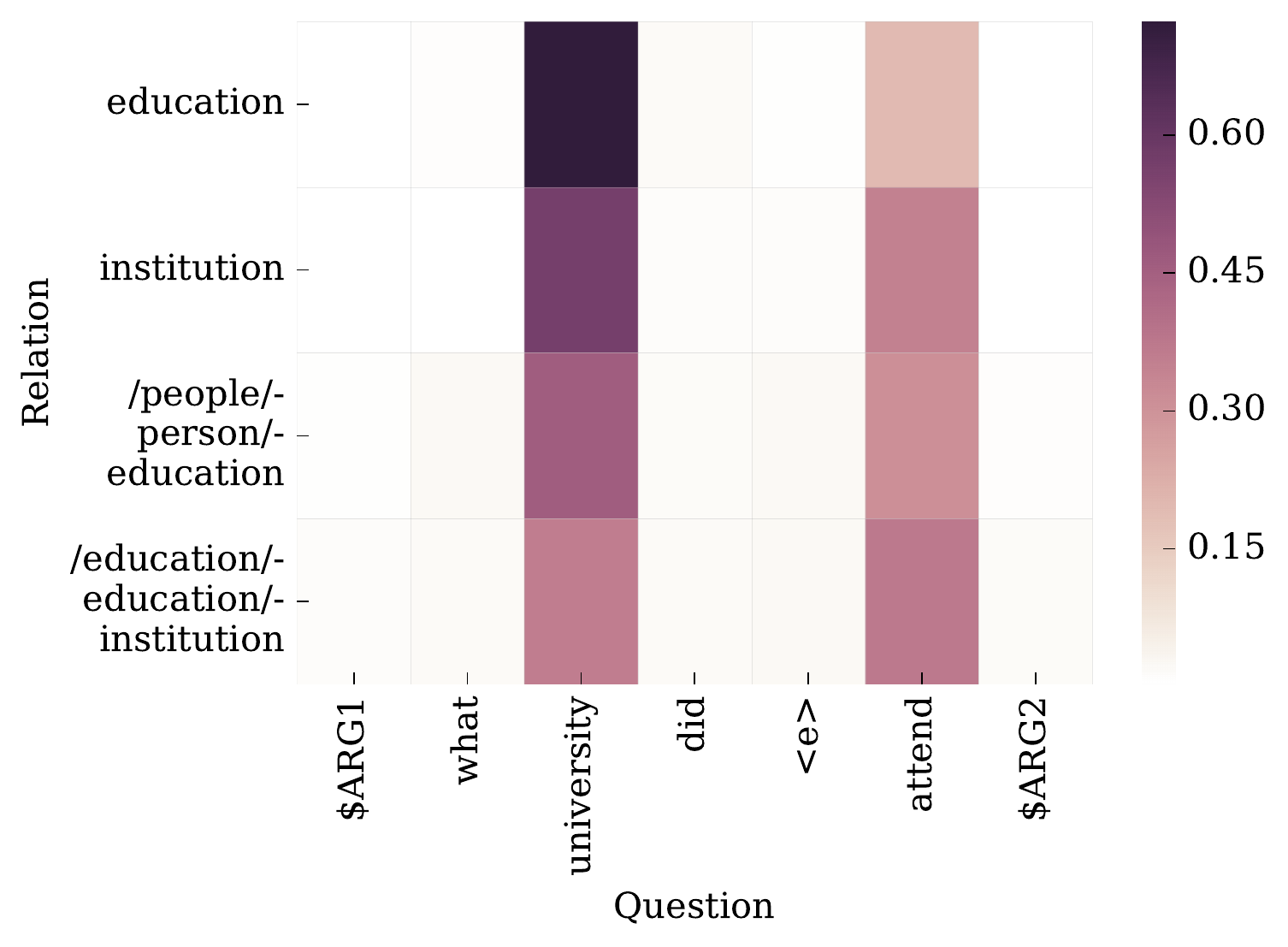}
		\caption{Heat-map of the attention weight matrix, which shows the soft alignment between the question (bottom) and relation (left).}
		\label{fig:Attention}
	\end{figure}
	
	\begin{figure}[!htb]
		\centering
		\includegraphics[width=0.45\textwidth]{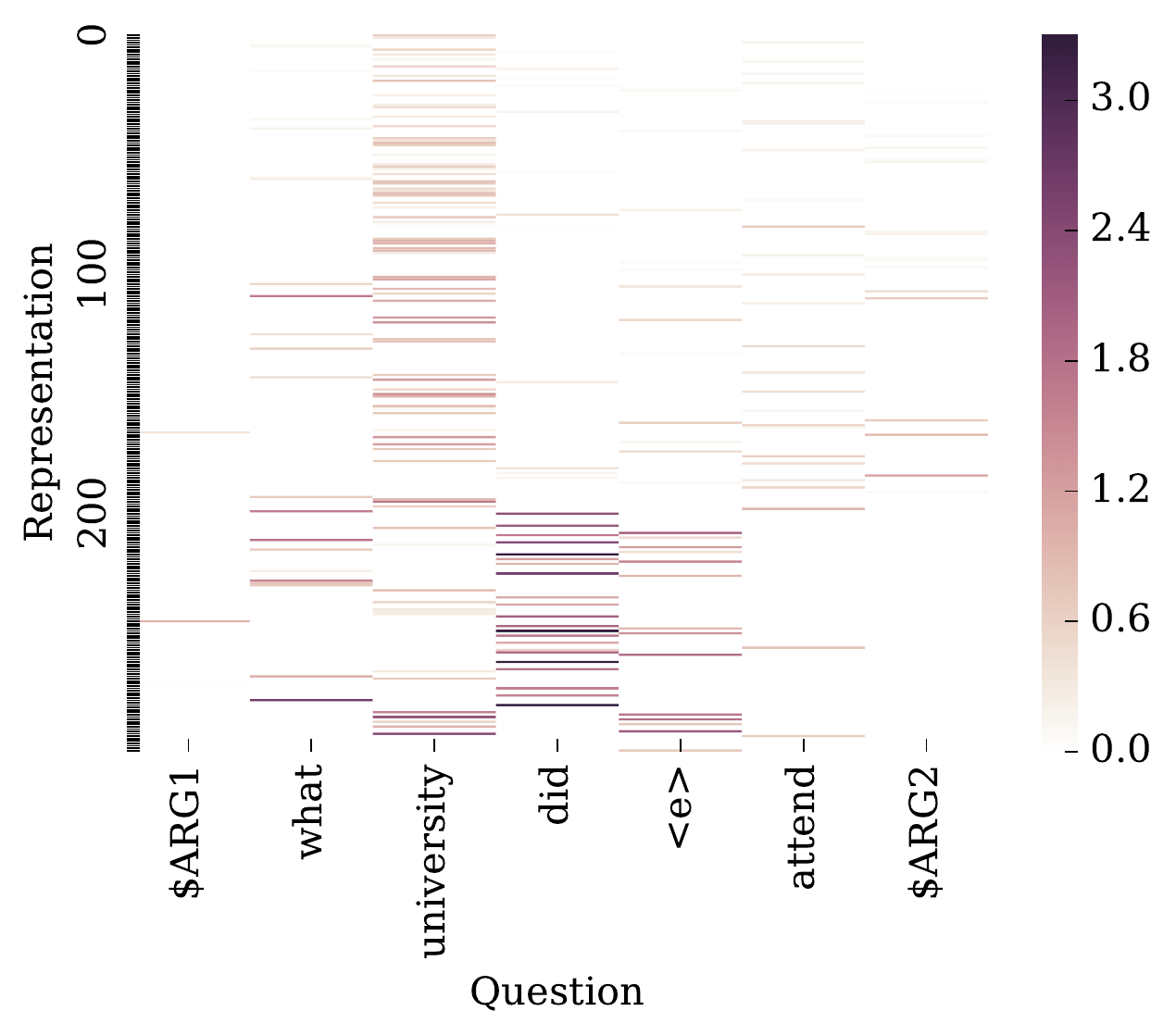}
		\caption{The visualization of the largest value of each dimension in the output of CNN layer. The first 100, second 100, and last 100 dimensions are generated with window size 1, 3 and 5 respectively.}
		\label{fig:maxvpos}
	\end{figure}
	
	The top of Table \ref{tab: abaltion_res} shows the ablation experiment results. 
	First, both tasks benefit from both the bi-directional attention mechanism, especially for WebQA dataset (84.44\% vs 78.41\%). Attention mechanism enables the model to focus on important parts of the sequence and get a better representation. 
	Second, comparing the sequences using word-level representations also outperforms the traditional encoding-comparing model, because multiple comparison operations are performed before information in the question is merged into a single vector. Finally, ABWIM, which combines the bi-directional attention and word-level interaction mechanisms, achieves state-of-the-art results on both tasks with accuracy 93.5\% and 85.32\% respectively. 
	
	Visualization of the attention matrix and the positions of maximum values obtained is given to further illustrate how the ABWIM works. 
	Visualization of the attention matrix learned is given in Figure \ref{fig:Attention}. The input question is ``what university did $\langle e\rangle$ attend'' and candidate relation path which contains two relation ``/people/person/education,  /education -/education/institution''. Depth of the colour represents the attention weights, {\color{black} and the darker the higher}.
	It shows that ABWIM learns a reasonable alignment between words in the question and the relation. For example,  more attention is paid to the words ``university'' and ``attend'', which are relevant to relation description. And it is also observed that the deep learning method is capable of capturing the semantic similarity between different symbols. 
	Visualization of the maximum value in each dimension of the CNN output is given in Figure \ref{fig:maxvpos}. It is observed that the largest values are obtained at the words which are crucial for relation detection. In other words, the CNN layer also learns to focus on important words when performing comparisons. 
	
	\subsubsection{Preprocessing layer}
	Ablation experiments are also carried out to study the effectiveness of the preprocessing Bi-LSTM layer which learns context aware representations. CNN and liner transformation layer are used to perform preprocessing, considering the difficulty in parallelization of the LSTM state computations. Specific ablation experiments are listed as follows. 
	\begin{itemize}
		\item Without preprocessing: simply removing the LSTM layers for question and relation preprocessing, and the word-level alignment and comparison are performed on their embeddings.
		\item Gated linear transformation: the question and relation are preprocessed as follows
		\begin{equation}
		\widehat{\bm{Q}}=\sigma\left(\bm{W}_i\bm{Q}\right)\odot\tanh\left(\bm{W}_u\bm{Q}\right)
		\end{equation}
		\begin{equation}
		\widehat{\bm{R}}=\sigma\left(\bm{W}_i\bm{R}\right)\odot\tanh\left(\bm{W}_u\bm{R}\right)
		\end{equation}
		where \(\bm{W}_i\in \mathbb{R}^{d_i\times d}\) and \(\bm{W}_u \in \mathbb{R}^{d_i\times d}\) are parameters to be learned, and \(d_i\) is a hyper-parameter.
		
		\item Fully CNN: unlike LSTM that depends on the computations of the previous timestep, CNN enables parallelization over every element in a sequence, so it is capable of making full use of the parallel architecture of GPU. We study the performance of fully CNN network on the relation detection of KBQA. The LSTM layer for question and relation preprocessing is replaced with a multi-kernel CNN layer, and the dimension of the CNN output is consistent with that of the original LSTM layer. 
		
	\end{itemize}
	
	Experimental results with different configurations on the context representation layer are given in the second parts of Table \ref{tab: abaltion_res}. First, ABWIM outperforms several strong baselines, though there is no preprocessing operation. It indicates the effectiveness of learning attention-based word alignments between question and relation. Second, a context aware representation of words can further improve the result and the preprocessing layer is also important for the overall performance. The LSTM layer, which is capable of learning long range dependency,　 outperforms the CNN and gated liner transformation. Finally, CNN and gated linear transformation do not rely on the computations of previous timestep, so they can fully utilize the computational capability of GPU and are more faster to be trained and perform inference. We leave the exploration of advanced CNN models with gated linear units \cite{gatedcnn} and residual connections \cite{residuallearning} as a future work.
	
	\section{Conclusion and future work}
	Relation detection is a key component of KBQA and it is difficult because of the flexible expressions of the relation in natural language and massive relations in the knowledge base. 
	In this paper, we propose to learn the attention-based word-level interactions between the question and the relation. Representations of relations are firstly mapped to the semantic space of every word in question with attention mechanism. Then multiple comparisons are performed on the word-level representations, and finally comparison results are merged to evaluate the semantic similarity. With multiple comparisons performed, ABWIM alleviates the bottleneck of compressing the question and the relation into fixed-dimensional vectors.  
	Experimental results of relation detection on both SimpleQuestions and WebQuestions show that the proposed model achieves state-of-the-art accuracy.
	
	In the future work, we will study transfer learning between relation detection of KBQA and the traditional relation extraction task, and further learn to utilize the great amount of training data generated by distant supervision methods to improve the performance of KBQA relation detection. 
	\section*{Acknowledgments}
	The authors thank Wenqiang Dong and Weili Zhang for the valuable comments on this work.
	\section*{References}
	
	\bibliography{mybibfile}
	
\end{document}